\documentclass[letterpaper]{article} % DO NOT CHANGE THIS
\usepackage[]{aaai24}  % DO NOT CHANGE THIS
\usepackage{times}  % DO NOT CHANGE THIS
\usepackage{helvet}  % DO NOT CHANGE THIS
\usepackage{courier}  % DO NOT CHANGE THIS
\usepackage[hyphens]{url}  % DO NOT CHANGE THIS
\usepackage{graphicx} % DO NOT CHANGE THIS

\usepackage{booktabs}
\usepackage{adjustbox}
\usepackage[numbers]{natbib}
\usepackage{diagbox}

\usepackage{natbib}  % DO NOT CHANGE THIS AND DO NOT ADD ANY OPTIONS TO IT
\usepackage{caption} % DO NOT CHANGE THIS AND DO NOT ADD ANY OPTIONS TO IT
\usepackage{algorithm}
\usepackage{algorithmic}

\usepackage{newfloat}
\usepackage{listings}
\DeclareCaptionStyle{ruled}{labelfont=normalfont,labelsep=colon,strut=off} % DO NOT CHANGE THIS
\floatstyle{ruled}
\newfloat{listing}{tb}{lst}{}
\floatname{listing}{Listing}
\title{RACCER: Towards Reachable and Certain Counterfactual Explanations for Reinforcement Learning}
\author {
    Jasmina Gajcin\textsuperscript{\rm 1},
    Ivana Dusparic\textsuperscript{\rm 1}
}

\affiliations {
    \textsuperscript{\rm 1}Trinity College Dublin\\
    gajcinj@tcd.ie, duspari@tcd.ie
}

\begin{document}

\maketitle

\begin{abstract}
While reinforcement learning (RL) algorithms have been successfully applied to numerous tasks, their reliance on neural networks makes their behavior difficult to understand and trust. Counterfactual explanations are human-friendly explanations that offer users actionable advice on how to alter the model inputs to achieve the desired output from a black-box system. However, current approaches to generating counterfactuals in RL ignore the stochastic and sequential nature of RL tasks and can produce counterfactuals that are difficult to obtain or do not deliver the desired outcome. In this work, we propose RACCER, the first RL-specific approach to generating counterfactual explanations for the behavior of RL agents. We first propose and implement a set of RL-specific counterfactual properties that ensure easily reachable counterfactuals with highly probable desired outcomes. We use a heuristic tree search of the agent's execution trajectories to find the most suitable counterfactuals based on the defined properties. We evaluate RACCER in two tasks as well as conduct a user study to show that RL-specific counterfactuals help users better understand agents' behavior compared to the current state-of-the-art approaches.
\end{abstract}

\section{Introduction}
Reinforcement learning (RL) has shown remarkable success in recent years and is being developed for high-risk areas such as healthcare and autonomous driving 
\cite{arulkumaran2017deep}. However, RL algorithms often use neural networks to represent their policies, making them difficult to understand and apply to real-life tasks \cite{puiutta2020explainable}.

Counterfactual explanations are user-friendly explanations for interpreting decisions of black-box algorithms \cite{miller2019explanation}. In machine learning, counterfactuals are defined as an answer to the question: \textit{``Given that the black-box model M outputs $A$ for input features $f_1, ...,f_k$, how can the features change to elicit output B from M?''} \cite{verma2021counterfactual}. They give users actionable advice on how to change their input to obtain a desired output, and are inherent to human reasoning, as we rely on them to assign blame and understand events \cite{byrne2019counterfactuals}.

In recent years, numerous methods for generating counterfactual explanations have been developed both for supervised \cite{wachter2017counterfactual,dandl2020multi,poyiadzi2020face,looveren2021interpretable,mothilal2020explaining,samoilescu2021model,laugel2017inverse,guidotti2018survey,guidotti2019factual} and RL \cite{olson2019counterfactual,huber2023ganterfactual}. In RL, \citet{olson2019counterfactual} propose a generative model for generating realistic counterfactuals that requires access to internal parameters of the black-box model. In contrast, \citet{huber2023ganterfactual} propose GANterfactual-RL, the only model-agnostic approach to generating counterfactuals in RL. GANterfactual-RL uses generative modeling to generate counterfactuals for visual tasks.

The majority of proposed methods for generating counterfactuals in supervised and RL search for the smallest change in features that leads to a change in outcome. However, due to the sequential nature of RL tasks, two states with similar features can be far away in terms of execution and even small changes in features can have uncertain 
outcomes due to stochasticity in the environment \cite{gajcin2022counterfactual}. Offering users counterfactuals that are not easy to reach or do not deliver on the promised outcome can cost users substantial time, and cause them to lose trust in the AI system. Additionally, current approaches do not distinguish between the two types of counterfactual explanations that can be defined for RL -- those that change causes from the past, from those that provide actionable advice for the future. For example, if a user's loan is denied, the counterfactual can either state that \textit{``Had your income been higher you would have been approved''}, or \textit{``If you increase your income, you will be approved in the future''} \cite{dai2022counterfactual}. 

In this work, we propose RACCER (\underline{R}eachable \underline{A}nd \underline{C}ertain \underline{C}ounterfactual \underline{E}xplanations for \underline{R}einforcement Learning), to the best of our knowledge the first approach for generating counterfactual explanations for RL tasks which takes into account the sequential and stochastic nature of the RL framework. RACCER generates explanations that explore how changes to the current state can affect future outcomes, sometimes referred to \textit{prefactual explanations} \cite{dai2022counterfactual}.  Firstly, we propose three novel RL-specific counterfactual properties -- \textit{reachability, stochastic certainty, and fidelity}. These counterfactual properties rely on the stochastic and sequential nature of RL tasks and ensure that counterfactuals are easy to reach and deliver the desired outcome with high probability. RACCER searches for the most suitable counterfactual by optimizing a loss function consisting of the three RL-specific properties using a heuristic tree search of the agent's execution tree. We evaluate RACCER in two environments and compare it to the only other model-agnostic approach for RL -- GANterfactual-RL \cite{huber2023ganterfactual}.  We find that RACCER performs better on feature-based and RL-specific counterfactual properties when explaining both fully-trained and suboptimal models. Additionally, we conduct a user study in which we compare the effect of counterfactual explanations on user understanding of RL agents and show that RACCER generates counterfactuals that help humans better understand and predict the behavior of RL agents.

Our contributions are as follows:

\begin{enumerate}
    \item We design three RL-specific counterfactual properties 
 -- reachability, stochastic certainty, and fidelity, and provide metrics for their estimation.   
    \item We propose RACCER, the first algorithm for generating RL-specific counterfactual explanations, which relies on the above counterfactual properties.
    \item We conduct a user study and show that RACCER can produce counterfactuals that help humans better understand an agent's behavior compared to the baseline approaches.
\end{enumerate}

The implementation of RACCER and evaluation details can be found at https://github.com/anonymous902109/RACCER.

\section{Related Work}

In supervised learning, counterfactual explanations have been used to propose changes in input features that elicit a desired prediction from a black-box model. Various counterfactual properties have been defined to evaluate different counterfactuals \cite{verma2021counterfactual}. For example, validity is used to measure whether counterfactual achieves the desired output, proximity is a feature-based similarity measure that ensures counterfactual features are similar to those in the original instance, and sparsity measures the number of features changed. In recent years, numerous works have proposed methods for generating counterfactual explanations in supervised learning \cite{wachter2017counterfactual,dandl2020multi,looveren2021interpretable,laugel2017inverse,mothilal2020explaining,poyiadzi2020face,samoilescu2021model}. The majority of these methods follow the same approach, where a loss function is defined by combining different counterfactual properties and optimized over the training data set. The methods differ in their design of the loss function and the choice of the
optimization method. For example, in the first work on counterfactual explanations for supervised learning, \citet{wachter2017counterfactual} use gradient descent to optimize a loss function based on proximity and validity properties. Similarly, \citet{mothilal2020explaining} propose DICE, which introduces a diversity property to the approach of \citet{wachter2017counterfactual} to ensure users are offered a set of diverse, high-quality explanations. \citet{dandl2020multi} poses the problem of counterfactual search as multi-objective optimization and uses a genetic algorithm to optimize validity, proximity, sparsity, and data manifold closeness of counterfactual instances. 

In RL, counterfactual explanations aim to explain a decision of a black-box RL model in a specific state by proposing an alternative state in which the model would choose the desired action. \citet{olson2019counterfactual} propose an approach that relies on generative modeling to create realistic counterfactuals, similar in features to the original instance, and produce a desired output. The approach is not model-agnostic and requires access to the internal parameters of the black-box model that is being explained. In contrast, \citet{huber2023ganterfactual} propose a model-agnostic approach GANterfactual-RL, which frames the counterfactual search as a domain translation problem, where each domain contains states in which the agent would choose a specific action. To find a suitable counterfactual, the original instance is translated to the target domain. The algorithm is based on the StarGAN architecture \cite{choi2018stargan} and includes training a discriminator $D$ and generator $G$. The generator receives as an input a state and target domain and produces a translated state. The role of the discriminator is to distinguish between real and fake images. The generator and discriminator are trained on states extracted from the agent's policy. 

Counterfactuals generated by current approaches in RL \cite{olson2019counterfactual,huber2023ganterfactual} generate realistic counterfactuals that can help users better understand agents' decisions and even detect faulty behavior in Atari agents. However, they focus on the same feature-based counterfactual properties such as proximity and sparsity as supervised learning methods. In RL where two states can be similar in features but distant in terms of execution, feature-based metrics are not sufficient for measuring how obtainable a counterfactual is \cite{gajcin2022counterfactual}. Relying only on feature-based similarity measures can produce counterfactuals that are not easily (or at all) obtainable, and decrease human trust in the system. In contrast, our work proposes the first approach for generating RL-specific counterfactuals that take into account the stochastic and sequential nature of RL tasks. 

Although the purpose of counterfactual explanations is to show a path to the desired outcome, this path can be uncertain due to the environment in which the system operates. For example, even if the loan applicant fulfills all conditions stipulated in a counterfactual, the bank might change the conditions for approving a loan. \citet{delaney2021uncertainty} recognize the need for estimating and presenting the uncertainty associated with counterfactuals to the user in supervised learning tasks. In this work, we estimate uncertainty from an RL perspective and use it not only as additional information for the user but as an important factor in searching for the counterfactual.

\section{RACCER}
\label{3}

In this section, we describe RACCER, our approach for generating counterfactual explanations for RL tasks. To generate a counterfactual explanation $x'$, we require oracle access to the black-box model $M$ being explained, the state $x$ being explained, and the desired outcome $a'$.  Additionally, the approach needs access to the RL environment. RACCER generates a counterfactual state $x'$ that can be easily reached from $x$ and in which the black-box model $M$ chooses $a'$ with a high probability. RACCER is fully model-agnostic, does not require information on model parameters, and can be used for generating counterfactual explanations of any RL model. 

RACCER does not search for the counterfactual directly but looks for a sequence of actions to transform the original into the counterfactual instance \cite{karimi2020algorithmic,ustun2019actionable,karimi2020survey}. This way of conducting counterfactual searches is more informative for the user, as they can be presented with not just the counterfactual instance, but also the sequence of actions they need to perform to obtain their desired outcome. To that end, we set out to find the optimal sequence of actions $A$ that can transform $x$ into a counterfactual state $x'$. In the remainder of this section, we describe how we can evaluate action sequences that lead to counterfactual states (Sections \ref{3.1} and \ref{3.2}) and describe our approach to searching for the optimal one (Section \ref{3.3}).

\subsection{Counterfactual Properties for RL}
\label{3.1}

Counterfactual properties guide the counterfactual search and are used to select the most suitable counterfactual explanation. In this section, we propose three RL-specific counterfactual properties that take into account the sequential and stochastic nature of RL tasks. These properties ensure that counterfactuals are easily obtainable from the original instance, and produce desired output with high certainty. We define these properties as functions of action sequence $A$ that transforms $x$ into counterfactual $x'$. 

\begin{figure*}[t]
    \centering
    \includegraphics[width=0.8\linewidth]{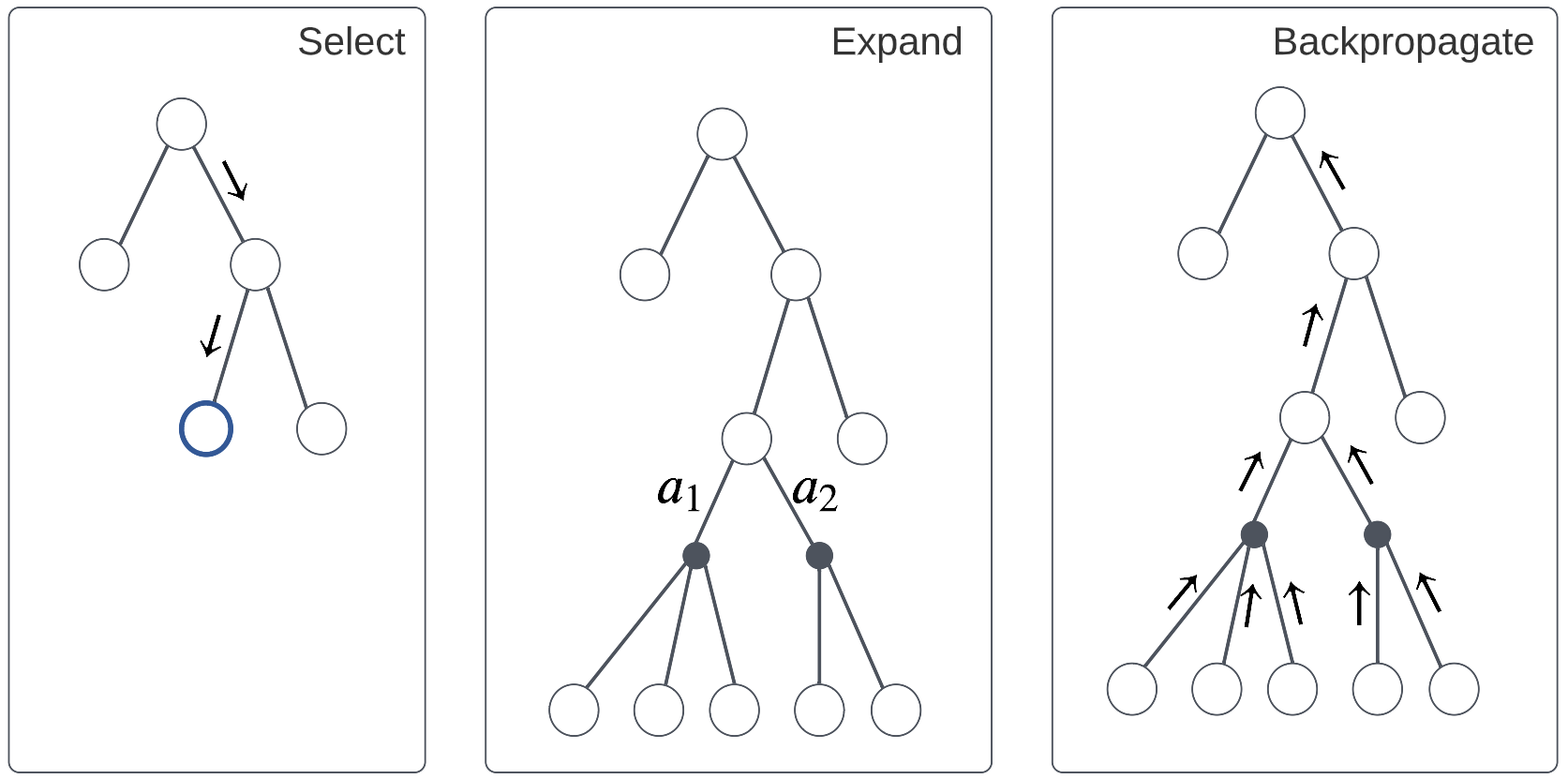}
    \caption{Heuristic tree search: in each iteration, a node is selected by navigating the tree from the root to a leaf by choosing actions according to the UCT formula. The node is expanded by performing all possible actions and appending all obtained states as children of the node. Finally, newly generated nodes are evaluated and their values are propagated back to the root to update the values of parent nodes. The white nodes represent states, while black nodes are determination nodes, that serve to instantiate all possible child states of a node in a stochastic environment. }
    \label{alg-fig}
\end{figure*}

\subsubsection{Reachability}

In RL two states can be similar in terms of state features, but far away in terms of execution. This means that, despite appearing similar, a large number of actions might be required to reach the counterfactual from the original state. Conversely, a state can be reachable by a few RL actions even if it appears different based on its feature values. Additionally, state features can be affected by stochastic processes outside of the agent's control. Relying solely on feature-based similarity measures (e.g. proximity, sparsity) could dismiss easily reachable counterfactuals where changes in features are beyond the agent's control and do not affect action choice.

To account for the sequential and stochastic nature of RL tasks, we propose measuring \textit{reachability}. For a state $x$ and a sequence of actions $A$, we define reachability as:

\begin{equation}
    R(x, A) = len(A)
\end{equation}

$R(x, A)$ measures the number of actions in the sequence that navigates to the counterfactual. Minimizing this property ensures counterfactuals can be reached within a small number of steps. 

\subsubsection{Fidelity}

RACCER searches for counterfactuals by finding an optimal sequence of RL actions to transform the original instance. For the counterfactual to be representative of the agent's behavior, the sequence of actions has to be likely under the agent's policy. As an example, consider a simple grid world where an agent needs to pick up one of the two keys -- red or blue and open the door. To explain why the agent did not choose to go to the door in a specific step, the counterfactual explanation might show that the agent would have gone to the door had they collected the red key first. However, if the agent's policy prefers the blue key over the red, this counterfactual is not representative of the agent's behavior and could be misleading to the user.

For this reason, RACCER prioritizes counterfactual states that can be reached under the agent's policy. We calculate the fidelity of a sequence of actions $A$ as the probability that the agent will choose these actions from state $x$:

\begin{equation}
    F(x, A) = 1 - \prod_{a \in A}
    {softmax(Q(x, \mathcal{A}))}[a]
    \label{fidelity}
\end{equation}

where $Q(x, a)$ is the Q-value of taking action $a$ in state $x$, and $\mathcal{A}$ is the action space of the task. By optimizing fidelity, we ensure that generated counterfactuals are representative of the agent's behavior.

\begin{algorithm}[t]
\caption{Counterfactual heuristic tree search}
    \begin{algorithmic}[1]
        \STATE \textbf{Input}: state $x$, desired outcome $a'$, black-box model $M$, environment $E$
        \STATE \textbf{Parameters}:number of iterations $T$
        \STATE \textbf{Output}: counterfactual state $x'$
        \STATE $t = \{x\}$ \hfill\COMMENT{Initializing search 
        tree}
        \STATE $i = 0$
        \WHILE{i < T}{
            \STATE n = select(t) \hfill\COMMENT{Select state $n$ to be expanded}
            \STATE S = expand(n) \hfill\COMMENT{Expand $n$ by performing available actions}
            \FORALL{$s \in S$}{ 
                \STATE $val(s) = L(x, A, a')$ \hfill\COMMENT{Evaluate states in $S$ according to $L$}
                \STATE $t += {s}$
            }\ENDFOR
            \STATE backpropagate() \hfill\COMMENT{Propagate values back to the root}
            \STATE $i += 1$
        }\ENDWHILE
        \STATE $p = []$
        \FORALL{$s \in t$} {
            \IF{$valid(s)$}{
                \STATE $p += s$ \hfill\COMMENT{Filter valid counterfactuals}
            }\ENDIF
        }\ENDFOR
        \STATE $cf = \arg \min_{s \in p} L(x, s(A), a')$ \hfill\COMMENT{Select the best counterfactual}
    \end{algorithmic}
    \label{tree_search}
\end{algorithm}

\subsubsection{Stochastic certainty}

One of the main qualities of counterfactual explanations is that they deliver the desired outcome. Asking the user to put their time and effort into changing the model inputs, only to obtain another unsatisfactory output can have detrimental effects on user trust in the system. During the time that is needed to convert the original instance into a counterfactual, the conditions of the task can change, rendering the counterfactual invalid.

In RL, the stochastic nature of the environment can make a counterfactual instance invalid during the time it takes to reach it from the original state. To ensure that users are presented with counterfactuals that are likely to produce the desired output, we propose \textit{stochastic certainty}. For instance $x$, a sequence of actions $A$, black-box model $M$ and the desired action $a'$ stochastic certainty is defined as:

\begin{equation}
    S(x, A, a') = 1 - P[M(x') = a' \quad|\quad x' = A(x)]
\end{equation}

where $A(x)$ is a state obtained by applying actions from $A$ to state $x$.
Intuitively, stochastic certainty measures the probability of the desired outcome still being chosen by $M$ after the time it takes to navigate to the counterfactual state. By maximizing stochastic certainty we promote sequences of actions that more often lead to the desired outcome.

\subsection{Loss Function} 
\label{3.2}
To optimize the counterfactual properties, we design a weighted loss function encompassing RL-specific objectives. For a state $x$, sequence of actions $A$, desired output $a'$, loss function is defined as:

\begin{equation}
    L(x, A, a') = \alpha R(x, A) + \beta F(x, A) + \gamma S(x, A, a')
    \label{L}
\end{equation}

where $\alpha, \beta$ and $\gamma$ are parameters determining the importance of different properties. By minimizing $L$ we can find a sequence of actions that quickly and certainly leads to a counterfactual explanation. However, $ L(x, A, a')$ does not verify that $a'$ is predicted in the obtained counterfactual. To that end, we additionally have to ensure that a validity constraint is satisfied:

\begin{equation}
    V(x, x', a') = M(x') == a'
    \label{validity}
\end{equation}

where $x'$ is obtained by performing actions from $A$ in $x$. Validity is used to filter potential counterfactual instances as is described in more detail in the next part of this section.

\subsection{Counterfactual Search}
\label{3.3}

Our goal is to obtain a sequence of actions $A$ that minimizes the loss function $L$ and satisfies the validity constraint.
Unlike traditional counterfactual search which directly searches for a counterfactual in a data set, we are looking for an optimal sequence of actions that can transform the original state into a counterfactual one. This means that we cannot directly optimize $L$ over a data set of states to find a counterfactual as this would give us no information about how difficult this counterfactual is to reach in terms of RL actions. To this end, we propose a counterfactual search algorithm that utilizes heuristic tree search to find a sequence of actions that transform the original into a counterfactual state that minimizes the loss function $L$. The details of the algorithm are given in Algorithm \ref{tree_search} and shown in Figure \ref{alg-fig}.

The proposed algorithm builds a tree to represent the agent's execution -- each node corresponds to a state, and each edge to one action. Each node $n$ is also associated with a value $val(n)$ and each edge is assigned a value $Q(n, a)$. These values are based on the loss function $L$ and are used to determine which node should be expanded in the next iteration. Children of a node are obtained by taking a specific action in that node. To account for the stochasticity in the environment, we apply determinization to the expanding process by adding hidden determinization nodes each time an action is performed. The children of determinization nodes are sampled from the possible states that result from performing a specific action. To calculate $val(n)$ we compute the value of $L(x, A, a')$, where $A$ is the sequence of actions that navigates from root $x$ to node $n$ in the tree. $Q(n, a)$ is calculated for each node $n$ and action $a$ as the average of values $val$ of the children nodes obtained when performing $a$ in $n$. To estimate $L(x, A, a')$ we need to calculate the values of individual counterfactual properties of reachability, fidelity, and stochastic uncertainty for nodes in the tree. We calculate the reachability of node $n$ as the length of the path between the root and $n$. Similarly, to calculate fidelity, we use the Q-values of state-action pairs on the path from the root to $n$ according to Equation \ref{fidelity}. Finally, to calculate stochastic certainty, we perform $N$ simulations by unrolling the sequence of actions $A$ from $x$ in the environment and record the number of times a desired outcome is obtained in the resulting state. We then calculate stochastic certainty as:

\begin{equation}
    S(x, A, a') = \frac{N(M(x') == a')}{N}
\end{equation}

where $x'$ is a state obtained after following $A$ in $x$. We normalize the values for reachability so that they fall within the $[0, 1]$ range, while fidelity and stochastic uncertainty values naturally belong to that range. We can then evaluate a node in a tree by combining and weighting the three counterfactual properties to obtain $L(x, A, a')$ as shown in Equation \ref{L}.

At the start of the search, a tree is constructed with only the root node corresponding to the state $x$ that is being explained. At each step of the algorithm, a node in the tree is chosen and the tree is expanded with the node's children. All actions are expanded simultaneously in the node. The resulting child nodes are then evaluated against $L$, and the results are propagated back to the tree root to update the value of nodes and edges. To decide which node is expanded in each iteration we navigate the tree from the root, at each node $n$ taking the action decided by the Upper Confidence Bound applied for Trees (UCT) formula \cite{kocsis2006bandit}:

\begin{equation}
    a^* = \arg \max_{a \in A} \left \{ Q(a, n) + C\sqrt{\frac{\ln(N(n))}{N(s,a)}} \right \}
\end{equation}

where $C$ is the exploration constant, $N(n)$ number of times $n$ was visited and $N(n, a)$ number of times $a$ was chosen in $n$. UCT balances between following the paths of high value and exploring underrepresented paths through the exploration constant $C$. The process is repeated until a predetermined maximum number of iterations $T$ is reached. 

Once the tree is fully grown, all nodes are first filtered according to the validity constraint (Equation \ref{validity}) to remain only with the states that deliver the desired output. The remaining nodes are potential counterfactual explanations. Then all nodes are evaluated against $L$. The state corresponding to the node in the tree with the minimum value for $L$ is presented to the user as the best counterfactual.

\section{Experiments}

In this section, we outline the experiment setup for evaluating RACCER. We compare RACCER to the only other model-agnostic algorithm for generating counterfactuals in RL -- GANterfactual-RL \cite{huber2023ganterfactual}. In Section \ref{4.1} we describe evaluation tasks.

\begin{table}[t]
    \centering
    \caption{Parameters used for generating counterfactual explanations for GANterfactual-RL and RACCER approaches in Stochastic GridWorld and Frozen Lake environments.}
    \begin{adjustbox}{width=\linewidth}
    \begin{tabular}{c|cc} \toprule 
        \backslashbox{Parameter}{Task} &Stochastic GridWorld & Frozen Lake\\ \midrule
        Number of iterations ($T$) & 300 & 300 \\
        Number of simulations ($N$) & 10 & 10\\
        Maximum number of actions ($k$) & 20 & 20\\
        Evaluation dataset size ($|D|$) & 500 & 400 \\
        Loss parameter $\alpha$ & -1 & -1\\
        Loss parameter $\beta$ & -1 & -1 \\
        Loss parameter $\gamma$ & -1 & -1\\
         \bottomrule    
    \end{tabular}
    \end{adjustbox}
    \label{param}
\end{table}

\begin{table*}[t]
    \centering
    \caption{The average values of counterfactual properties for counterfactual explanations of a fully-trained agent generated using GANterfactual-RL and $RACCER$ approaches in Stochastic GridWorld and Frozen Lake.}
    \begin{adjustbox}{width=1\linewidth}
    \begin{tabular}{c|cc|cc} \toprule 
        Task & \multicolumn{2}{c|}{Stochastic Gridworld} & \multicolumn{2}{c}{Frozen Lake} \\ \midrule
        \backslashbox{Metric}{Approach} & GANterfactual-RL & RACCER & GANterfactual-RL & RACCER\\ \midrule
        Generated counterfactuals ($\%$) & \textbf{100} & 75.4 & \textbf{100} & 80.75 \\
        Realistic counterfactuals ($\%$) & 76.0 & \textbf{100} & \textbf{100} & \textbf{100}\\ \midrule
        Proximity ($\uparrow$) & 0.98 &\textbf{ 0.99 }& 0.90 & \textbf{0.96}\\
        Sparsity ($\downarrow$)  & 0.19 & \textbf{0.11} & 0.61 & \textbf{0.14}\\
        Validity ($\uparrow$) & 0.58 & \textbf{1.0} & 0.46 & \textbf{1.0}\\ \midrule
         Reachability ($\downarrow$)& 0.98 &\textbf{0.13} & 1.0 & \textbf{0.15} \\
        Fidelity ($\downarrow$) & 1.0 & \textbf{0.79} & 1.0 & \textbf{0.6}\\
        Stochastic uncertainty ($\downarrow$) & 0.99 &\textbf{ 0.18} & 1.0 & \textbf{0.08} \\
        \bottomrule    
    \end{tabular}
    \end{adjustbox}
    \label{results}
\end{table*}

\subsection{Evaluation Tasks}
\label{4.1}

We evaluate our approach in two environments -- Stochastic GridWorld and Frozen Lake.

\subsubsection{Stochastic GridWorld}

Stochastic GridWorld is a simple $5 \times 5$ grid world, where the agent is tasked with shooting the dragon. To successfully shoot the dragon, the agent has to be in the same file or row as the dragon, and the space between them has to be empty. In that situation, the agent can successfully perform the SHOOT action and win the game. The environment also contains trees and walls in the middle file of the grid, that can block the agent's path to the dragon. An agent can chop down a tree or a wall by performing a CHOP action when located directly near it. However, trees are less costly to chop down than the walls. At each step, the agent can move one step in any direction or perform SHOOT and CHOP actions. Additionally, along the middle file of the board, trees can regrow and walls can be rebuilt with different probabilities. Actions receive a $-1$ penalty, and successfully shooting the dragon brings a $+10$ reward. The episode ends when the dragon is shot or when the maximum number of time steps is reached. We consider all states that contain an agent, a dragon, and have trees and walls only along the middle file of the grid to be realistic under the game rules.

In this task, two states can appear similar but be far away in terms of execution, due to the obstacles on the grid. Chopping down a tree to be able to shoot the dragon might be less preferable than going around it, and suggesting this to the user could save them time and effort. Similarly, due to the stochastic nature of the task, during the time needed to obtain a counterfactual, new trees and walls can regrow and block the agent's path to the dragon. 

\subsubsection{Frozen Lake}

The frozen lake is a well-known stochastic grid world environment, in which the agent is tasked with reaching the goal while navigating a grid where some squares are covered in ice. Making an action in icy states can either lead the agent to a desired state or leave them in the same one with some probability. All actions carry a -1 reward, while successfully navigating to the goal brings the agent +10. We consider all states that contain an agent and the goal to be realistic under the rules of the game.

In this environment, two states with very similar features can be far away. For example, even if there is only one square difference between the agent's locations in two states, it might still be difficult to reach one from the other given the stochastic nature of the environment. 

\section{Evaluation}
\label{5}

To evaluate RACCER, we compare it to the baseline approach GANterfactual-RL \cite{huber2023ganterfactual} in Stochastic GridWorld and Frozen Lake environments. We establish 5 hypotheses for evaluating RACCER:

\begin{itemize}
    \item \textbf{H1:} Counterfactual explanations generated by RACCER will perform better on RL-specific metrics of fidelity, stochastic uncertainty, and reachability compared to the baseline.
    \item \textbf{H2:} RACCER is more suitable for producing counterfactuals for explaining suboptimal agents compared to the baseline.
    \item \textbf{H3:} RACCER will produce counterfactuals that can help users better understand and predict the behavior of RL agents compared to the baseline.
    \item \textbf{H4:} RACCER will produce counterfactuals that help users better choose between agents with different preferences compared to the baseline. 
    \item \textbf{H5:} Counterfactual explanations generated by RACCER will be perceived as more satisfactory by users compared to explanations generated by the baseline. 
\end{itemize}

To evaluate H1 and H2, we evaluate the counterfactual properties of explanations generated by RACCER and GANterfactual-RL. Hypothesis H1 is explored in Section \ref{5.1.1}, and H2 in Section \ref{5.1.2}. Hypotheses, H3, H4 and H5 are evaluated through a user study described in detail in Section \ref{5.2}. Hypothesis H3 is evaluated in Section \ref{5.2.1}, H4 in Section \ref{5.2.2} and H5 in Section \ref{5.2.3}.

\subsection{Evaluating Counterfactual Properties}
\label{5.1}

We evaluate RACCER and GANterfactual-RL based on both feature-based counterfactual properties (proximity, sparsity, validity, and realistic counterfactual) and RL-specific properties (reachability, fidelity, stochastic uncertainty).

To evaluate proximity, sparsity, and validity we use metrics defined in \citet{huber2023ganterfactual} originally used to evaluate GANterfactual-RL. For proximity, we use the L1 distance between instances:

\begin{equation}
    P(x, x') = 1 - ||x - x'||_{1}
\end{equation}

Sparsity is calculated as the number of non-modified features when transforming the original instance $x$ into a counterfactual $x'$:

\begin{equation}
    S(s, s') = \frac{||x - x'||_{0}}{S}
\end{equation}

where $S$ is the total number of features. 

Validity denotes whether the target action $a'$ is chosen by the black-box model $B$ in the counterfactual instance $x'$:

\begin{equation}
    V(x') = B(x') == a'
\end{equation}

Additionally, we also evaluate whether the resulting counterfactuals are realistic. What constitutes a realistic counterfactual is task-specific and is described in more detail in Section \ref{4.1}.

Furthermore, we evaluate RACCER and GANterfactual-RL according to RL-specific properties presented in Section \ref{3.1}. Evaluating these properties for RACCER is straightforward as it uses tree search to navigate to the counterfactual. Properties can be calculated by analyzing the sequence of actions leading from the root to the counterfactual. GANterfactual-RL, however, generates counterfactual using generative models and uses no notion of actions. To measure reachability, fidelity, and stochastic certainty for a counterfactual $x'$ generated by GANterfactual-RL, we build a tree of the agent's execution of length $k$ rooted in $x$ and find $x'$ in it. In that way, we can estimate properties that rely on actions even for explanations generated through the direct search for counterfactual states. If $x'$ cannot be found in the tree, it is assigned the least desirable value for an RL-specific counterfactual property which is $1$.

\begin{table*}[t]
    \centering
    \caption{The average values of counterfactual properties for counterfactual explanations for a sub-optimal agent generated using GANterfactual-RL and $RACCER$ approaches in Stochastic GridWorld and Frozen Lake for a suboptimal agent $M_{sub}$.}
    \begin{adjustbox}{width=1\linewidth}
    \begin{tabular}{c|cc|cc} \toprule 
        Task & \multicolumn{2}{c|}{Stochastic Gridworld} & \multicolumn{2}{c}{Frozen Lake} \\ \midrule
        \backslashbox{Metric}{Approach} & GANterfactual-RL & RACCER & GANterfactual-RL & RACCER\\ \midrule
        Generated counterfactuals ($\%$) & \textbf{100} & 77.6 & \textbf{100} & 43.50 \\
        Realistic counterfactuals ($\%$) & 47.00 & \textbf{100} & \textbf{100} & \textbf{100}\\ \midrule
        Proximity ($\uparrow$)& 0.98 &\textbf{ 0.99 }& 0.87 & \textbf{0.97}\\
        Sparsity ($\downarrow$)  & 0.24 & \textbf{0.11} & 0.81 & \textbf{0.14}\\
        Validity ($\uparrow$) & 0.55 & \textbf{1.0} & 0.2 & \textbf{1.0}\\ \midrule
         Reachability ($\downarrow$)  & 0.99 &\textbf{0.14} & 1.0 & \textbf{0.10} \\
        Fidelity ($\downarrow$) & 1.0 & \textbf{0.82} & 1.0 & \textbf{0.76}\\
        Stochastic uncertainty ($\downarrow$) & 1.0 &\textbf{0.26} & 1.0 & \textbf{0.07}  \\
        \bottomrule    
    \end{tabular}
    \end{adjustbox}
    \label{results_suboptim}
\end{table*}

\subsubsection{Explaining Fully-trained Agents}
\label{5.1.1}

We start by comparing explanations generated by RACCER and GANterfactual-RL when explaining a fully trained agent. We obtain a fully trained black-box model $M$ by training a DQN on the task until convergence. We then apply RACCER and GANterfactual-RL approaches to generate counterfactuals for explaining $M$. The parameters for generating counterfactuals using both algorithms are given in Table \ref{param}.
Both RL-specific and feature-based counterfactual properties are evaluated for the generated counterfactuals. We also record what percentage of counterfactuals were successfully created and if they were realistic. The results for both tasks are recorded in Table \ref{results}.

\textbf{In both environments, RACCER performs better on both feature-based and RL-specific counterfactual metrics.} While we expected RACCER to perform better on RL-specific properties, it is surprising that it outperforms GANterfactual-RL in feature-based metrics, as GANterfactual-RL has been trained to optimize these. We speculate that this is because the GANterfactual-RL approach has been optimized for visual tasks, unlike discrete environments used in this work. RACCER also produces only realistic counterfactuals as it follows the rules of the environment. GANterfactual-RL, on the other hand, often changes features outside of the agent's control such as adding or removing tree and wall features in the Stochastic GridWorld environment, resulting in fewer realistic states. Finally, RACCER generates counterfactuals that are more often valid.

One metric in which RACCER performs worse compared to the baseline is the number of generated counterfactuals. Due to its underlying generative model, GANterfactual-RL can generate a counterfactual for each fact and target action. RACCER, on the other hand, searches the space of the agent's interactions with the environment to find a counterfactual. If the agent is very unlikely to play a certain action in the environment, RACCER will not be able to generate a counterfactual for this action. For example, if the monster in the Stochastic GridWorld environment is located in the rightmost file of the grid, a well-trained agent will never need to play action LEFT. By examining the factual states and target actions for which RACCER does not generate a counterfactual we find that a large majority of them correspond to states where the target action would never be played by a well-trained agent. Specifically, in Stochastic GridWorld out of 123 situations where RACCER does not generate a  counterfactual, in 80 of them ($65.04\%$) target action would never be played by an agent. This means that RACCER fails to find a counterfactual in a situation where that is possible only 43 times, or for $8.6\%$ of situations. Similarly, in Frozen Lake, out of $77$ situations in which RACCER does not find a counterfactual, $72$ corresponds to such impossible situations. In the Frozen Lake task, RACCER fails to generate counterfactuals where that is possible only 5 times, or for $0.0125\%$ situations.

\begin{figure*}[t]
    \centering
    \includegraphics[width=\linewidth]{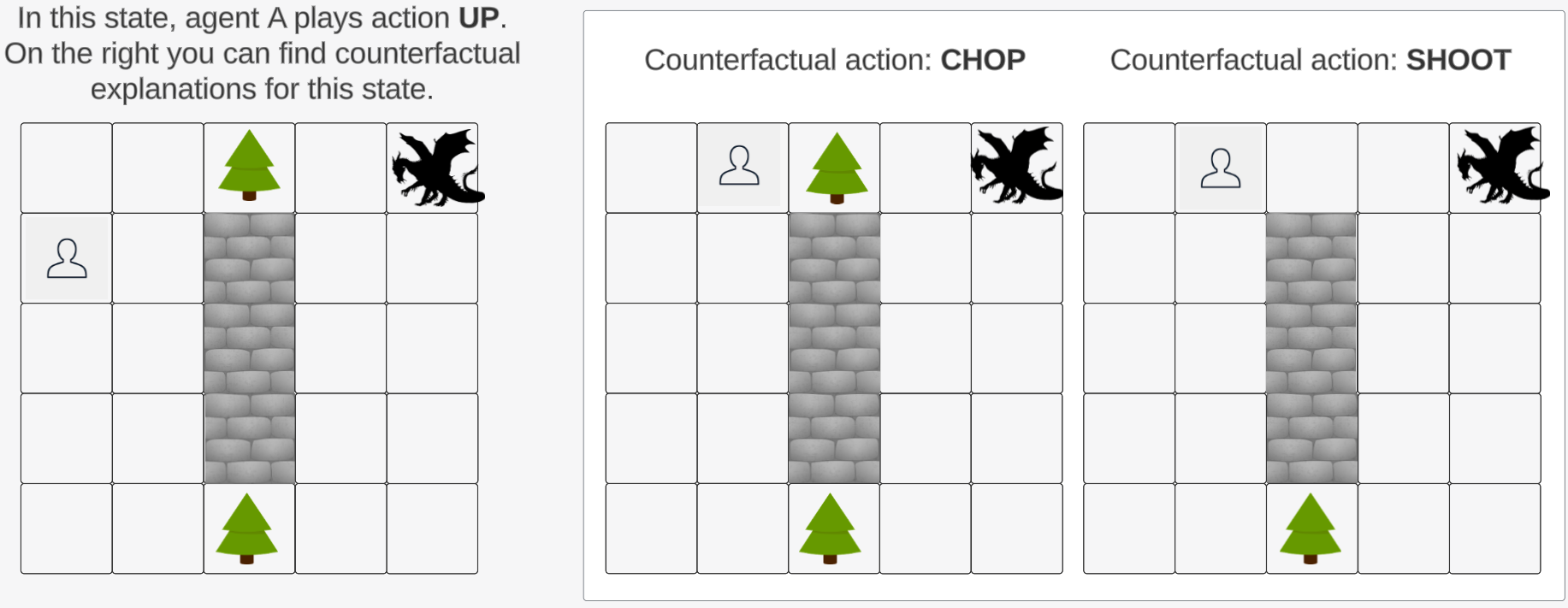}
    \caption{Example of the counterfactual explanation shown to the user during the training phase of the user study.}
    \label{user_study_example}
\end{figure*}

\subsubsection{Explaining Suboptimal Agents}
\label{5.1.2}

Most often, counterfactual explanations have been applied to explain the behavior of fully-trained agents. However, understanding suboptimal agents is necessary for verification and debugging. For example, \citet{olson2019counterfactual} have used counterfactuals to help users recognize agents that relied on artificially inserted pixels correlated with an action choice. Agents relying on these spurious correlations do not base decisions on actual game elements, and their performance suffers when the spurious correlation is broken. 

GANterfactual-RL relies on training supervised learning models to translate states between domains. We hypothesize that this approach, although suitable for explaining fully-trained agents, cannot be applied to suboptimal ones. This is because domains for training the generator and discriminator models in GANterfactual-RL are defined based on which actions the RL agent would make in a specific state. However, if an agent is not fully-trained there is likely to be some randomness in its decision-making process. This introduces randomness into domains, resulting in domains that are difficult to separate, making supervised learning of discriminator and generator models challenging.

To train a suboptimal agent $M_{sub}$ in both tasks, we use a DQN model, but train it for one-tenth of the time used to train the fully-trained agent. We evaluate RACCER and GANterfactual-RL on proximity, sparsity, and validity, as well as reachability, fidelity, and stochastic uncertainty. Additionally, we record the percentage of generated counterfactuals as well as the percentage of realistic counterfactuals. The results are presented in Table \ref{results_suboptim}.

\textbf{RACCER performs comparably when explaining a suboptimal agent $M_{sub}$ (Table \ref{results_suboptim}) and the fully-trained agent $M$ (Table \ref{results}) in both environments}. In contrast, counterfactuals generated by GANterfactual-RL show a decline in counterfactual properties when explaining a suboptimal model compared to a fully trained model. In the Frozen Lake environment, GANterfactual-RL achieves lower values for validity when explaining a suboptimal model compared to a fully-trained model. Similarly, in the Stochastic Gridworld task, GANterfactual-RL generates counterfactuals that are far less realistic compared to those generated for a fully-trained model.

\subsection{User Study}
\label{5.2}

Counterfactual explanations are ultimately intended to assist humans in real-life tasks, and evaluating them in this context is necessary to ensure their usefulness. To evaluate hypotheses H3, H4, and H5 we conducted a user study to compare the counterfactual explanations produced by GANterfactual-RL and RACCER. We conducted the study in the Stochastic GridWorld environment, as it has simple rules, and requires no prior knowledge from users.

We sourced 153 participants through the Prolific platform from English-speaking countries (UK, Ireland, Canada, USA, Australia, and New Zealand) and split them into two groups. The first group received counterfactuals generated by GANterfactual-RL and the second counterfactuals produced by RACCER. After filtering participants for those who had passed attention checks, we remained with 58 participants in the first and 63 in the second group. Participants were remunerated for their time according to the Prolific payment policy.

The study consisted of 3 parts -- evaluating user understanding of the agent's behavior, evaluating user understanding of the agent's preferences, and evaluating user satisfaction. The study design follows that used to evaluate the GANterfactual-RL algorithm in \citet{huber2023ganterfactual}. Before starting the study, users were presented with general information about the task and the study. Users were also shown a definition and examples of counterfactual explanations and asked to answer test questions to ensure a full understanding of the task. The template for the study can be found at: https://qrxhyre44mt.typeform.com/to/cpeLrWbZ. Section \ref{5.2.1} covers the evaluation of agent's behavior, Section \ref{5.2.2} understanding of agents' preferences, and Section \ref{5.2.3} user satisfaction with counterfactual explanations.

\subsubsection{Agent Understanding}
\label{5.2.1}
To evaluate how well users understand agent's behavior we use a user study setup similar to that of \citet{huber2023ganterfactual}. Users are shown the behavior of two agents A and B with different policies, described in more detail in Section \ref{5.2.2}. For each agent users go through two stages -- training and testing stage. In the training stage, users are shown a game state and the action agent chooses that state. Then, users are presented with counterfactual states, describing in which situations the agent would choose an alternative action. For each agent, the user sees 10 factual states during the training phase. Figure \ref{user_study_example} shows an example explanation shown to users during the training phase. In the testing phase, users are presented with $10$ states and asked to predict an action the agent would take, without being given the explanations. 

The factual states in the training and testing phase are selected by the HIGHLIGHTS-DIV algorithm \cite{amir2018highlights}, inspired by the setup from \citet{huber2023ganterfactual}. This way users are presented with the most informative states of the agent's game-play. We modify the HIGHLIGHTS-DIV algorithm to include states with a diverse range of Q-values since using the original HIGHLIGHTS-DIV algorithm results in a mostly homogeneous set of states in which the agent should perform the SHOOT action. We generate 20 most informative states according to HIGHLIGHTS-DIV and randomly split them into training and testing sets. To be able to show counterfactuals to the users, they need to be realistic. For that reason, we additionally filter the states obtained by the HIGHLIGHTS-DIV algorithm to ensure they are realistic. We present two counterfactual states for each factual one, to reduce the cognitive load required by the experiment. We show counterfactuals for actions CHOP and SHOOT, as these actions represent the most interesting game-play. 

We use the prediction accuracy of the agent's actions in the testing phase as a metric for measuring user understanding of the agent's behavior. Users who have seen counterfactuals generated by RACCER have shown $76.19\%$ accuracy in predicting agents' actions. In contrast, users who have been presented with counterfactuals generated using the GANterfactual-RL approach have achieved an accuracy of $70.94\%$. After conducting a non-parametric one-tailed Mann-Whitney U test we find a significant difference between the prediction accuracy of the two approaches (p $= 0.0185$). This proves H3 and \textbf{indicates that RL-specific counterfactuals help users better understand and predict the behavior of RL agents.}

\subsubsection{Agent Comparison}
\label{5.2.2}

During the user study, users were asked to evaluate two agents A and B one after the other. Agents A and B are both fully trained on the task. However, they are trained on different reward functions and have different preferences for completing the task. Specifically, Agent A prefers to take the longer, but cheaper paths in the environment, while Agent B does not care about the cost and wants to finish the task as quickly as possible. The difference in the behavior between the two agents is exhibited most clearly in their interaction with the wall features. When faced with a wall obstacle, Agent A chooses to go around it to reduce costs, while Agent B chooses to chop down the wall despite the high cost to finish the task quicker. 

Each user is first presented with the training and testing phase for one agent, followed by the training and testing phase for the second agent. The users are informed when they will be switching from one agent to the other. After finishing the training and testing phases for both agents, users are asked to choose a more suitable agent according to a specific preference. Specifically, users are asked which agent they would choose if they wanted to keep the cost minimal, regardless of the time it takes to finish the task. Conversely, they were also asked which agent would they choose to finish the task quickly, regardless of the cost. 

Users presented with RACCER explanations choose the correct agent in $53.17\%$, while users who have seen GANterfactual-RL explanations made a correct choice in $58.62$ of cases (p = $0.6509$). This indicates that \textbf{contrary to H4, RACCER is not better at helping users distinguish between agents with different policies compared to GANterfactual-RL. }

\subsubsection{User Satisfaction}
\label{5.2.3}
At the end of the study, users were also asked to rank the explanations based on the \textit{explanation goodness metrics} \cite{hoffman2018metrics} on a $1-5$ Likert scale ($1$ - strong disagreement, $5$ - strong agreement). Specifically, users reported whether they found explanations to be useful, satisfying, complete, detailed, actionable, trustworthy, and reliable. 

\begin{figure}[t]
    \centering
    \includegraphics[width=\linewidth]{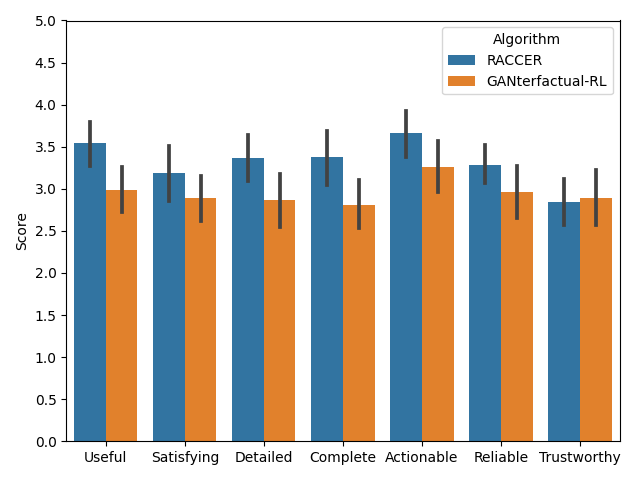}
    \caption{Users' scores on explanation goodness metrics \cite{hoffman2018metrics} for counterfactual explanations generated using RACCER and GANterfactual-RL algorithms.}
    \label{scores}
\end{figure}

The results of this part of the study are presented in Figure \ref{scores}. After conducting 
a non-parametric one-tailed Mann-Whitney U test we find that users perceive explanations generated by \textbf{RACCER to be significantly more useful for understanding the agent (p = 0.0057), more detailed (p = 0.0190) and complete (p = 0.0095) compared to those generated by GANterfactual-RL approach.} However, there is no significant difference between the approaches in the perceived trustworthiness (p = 0.7901), reliability (p = 0.1446), and actionability (p = 0.0729) of explanations, resulting in H5 being only partially confirmed by our experiments.

\section{Conclusion and Future Work}

In this work, we presented RACCER, the first RL-specific approach to generating counterfactual explanations. We designed and implemented three novel counterfactual properties that reflect the sequential and stochastic nature of RL tasks, and provided a heuristic tree search approach for optimizing these properties. We evaluated our approach in Stochastic GridWorld and Frozen Lake environments and showed that RACCER generates counterfactuals that are easier to reach and provide the desired outcomes more often compared to baseline approaches. We have also conducted a user study, and shown that RACCER helps users better predict the behavior of RL agents, and produces explanations that are perceived as more useful, detailed,d and complete compared to GANterfactual-RL. 

In this work, we have limited our search to only the best counterfactual. In future work, we hope to expand our search to include a set of diverse counterfactual explanations optimizing different counterfactual properties. In this way, users would have a wider choice of potentially actionable advice. Additionally, we have only explored the prefactual explanations which explore how changes in the current state can lead to different outcomes. In future work, we hope to investigate counterfactuals that explore past decisions and compare them to prefactuals in RL.

\bibliography{aaai24}

\end{document}